\documentclass{article} 

\usepackage[english]{babel}

\usepackage[utf8]{inputenc}
\usepackage[margin=1.5in]{geometry}
\usepackage{amsmath}
\usepackage{amsthm}
\usepackage{amsfonts}
\usepackage{amssymb}
\usepackage[usenames,dvipsnames]{xcolor}
\usepackage{graphicx}
\usepackage[siunitx]{circuitikz}
\usepackage{tikz}
\usepackage[colorinlistoftodos, color=orange!50]{todonotes}
\usepackage{hyperref}
\usepackage{natbib}
\usepackage{har2nat} 
\usepackage{fancybox}
\usepackage{epsfig}
\usepackage{soul}
\usepackage[framemethod=tikz]{mdframed}

\usepackage[onehalfspacing]{setspace}
\usepackage{url}
\usepackage{breakurl} 

\usepackage{microtype}

\newcounter{points}
\newcounter{spelling}
\newcounter{usage}
\newcounter{units}
\newcounter{other}
\newcounter{source}
\newcounter{concept}
\newcounter{missing}
\newcounter{math}

\title{
\normalfont \normalsize 
\rule{\linewidth}{0.5pt} \\[6pt] 
\huge The Virtuous Machine - \\ Old Ethics for New Technology?\\
\rule{\linewidth}{2pt}  \\[1pt]
}

\begin{document}

\maketitle
\begin{center}
{\large \centering Nicolas Berberich\textsuperscript{*,1,2,3}, Klaus Diepold\textsuperscript{1,2}}

  \vspace{5mm}
  {\textsuperscript{1} Department of Electrical and Computer Engineering, Technical University of Munich\\
  \textsuperscript{2} Department of Informatics, Technical University of Munich\\
  \textsuperscript{3} Munich Center for Technology in Society \\
  \textsuperscript{*} E-mail: n.berberich@tum.de
  }
\end{center}

\noindent
\text{}



\begin{abstract} 
Modern AI and robotic systems are characterized by a high and ever-increasing level of autonomy. At the same time, their applications in fields such as autonomous driving, service robotics and digital personal assistants move closer to humans. From the combination of both developments emerges the field of AI ethics which recognizes that the actions of autonomous machines entail moral dimensions and tries to answer the question of how we can build moral machines.

In this paper we argue for taking inspiration from Aristotelian virtue ethics by showing that it forms a suitable combination with modern AI due to its focus on learning from experience. 

We furthermore propose that imitation learning from moral exemplars, a central concept in virtue ethics, can solve the \textit{value alignment problem}. 
Finally, we show that an intelligent system endowed with the virtues of temperance and friendship to humans would not pose a \textit{control problem} as it would not have the desire for limitless self-improvement.  
\end{abstract}

\section{Introduction}


Modern technological systems exhibit an ever increasing degree of autonomy, due to rapid  progress in the fields of robotics and artificial intelligence. Familiar examples are autonomous vehicles, care robots and intelligent weapon systems. 
Like every action, the actions of these artificial cognitive systems involve moral dimensions, with the peculiarity that they can't be traced back to the developer, producer or user, which was possible with traditional technological artifacts. Due to the inherent autonomy of these systems, the ethical considerations have to be conducted by themselves. This means, that these autonomous cognitive machines are in need of a theory, with the help of which they can, in a specific situation, choose the action that adheres best to the moral standards. 
Whether the machine then possesses true morality or if it only simulates it will not be discussed here. We use the term 'moral' in the same functional way in which 'intelligence' is used in AI research. If the moral behavior of the machine imitates that of a human well enough that an interaction partner can't distinguish them, we can, for all practical purposes, say that the machine is moral. This moral Turing test might be used as evaluation criterion for any asserted moral machine.  
Of central importance is acceptance. Contrary to common interpretations, Masahiro Mori's concept of the \textit{uncanny valley} \cite{mori1970uncanny} is not only restricted towards looks, but also the behavior of human-like machines. It claims that humans develop an unconscious aversion against machines, when these look similar to them, but not quite the same. 

A psychological explanation for this phenomenon is that a cognitive dissonance arises from the mismatch of what one expects a human to look like and what the android actually looks like. When the visual appearance of a human is similarly off, this can be a sign of a disease, so the cognitive dissonance of the uncanny valley might be explained as a evolutionarily evolved survival mechanism, an argument brought forward by Mori himself. An even stronger cognitive dissonance takes place when there is a mismatch between the looks and the behavior of an artificial agent. 
As MacDorman and Chattopadhyay argue, not the category uncertainty between human and non-human is the cause for the cold and eerie feelings of the uncanny valley, but realism inconsistency \cite{macdorman2016reducing}. This is why there is no uncanny valley effect in industrial robots or vacuum cleaning robots - they are consistent in how they look and what they do and are thus on the left side of the valley. When cognitive systems become more autonomous and this allows their application domains to rapidly move closer to humans (from production floors to living rooms), their behavior, including moral actions, might not keep pace. This discrepancy between what people believe that technology can do, based on its appearance, and what it actually can do, would not only elicit a strong uncanny valley effect, but also pose a large safety risk. 
Taken together, we predict that this would lead to an acceptance problem of the technology. 
If we want to avoid this by jumping over the uncanny valley, we have to start today by thinking about how to endow autonomous cognitive systems with more human-like behavior. 
The position that we argue for in this paper is that the last discrepancy between the valley and its right shore lies in virtuous moral behavior. In the near future we will have autonomous cognitive machines whose actions will be akin to human actions, but without consideration of moral implications they will never be quite alike, leading to cognitive dissonance and rejection. We believe that taking virtue ethics as the guiding moral theory for building moral machines is a promising approach to avoid the uncanny valley and to induce acceptance.

Besides the social agreeability of autonomous cognitive systems, their impact on society as a whole has become an object of debate in recent years. Social issues like impending mass unemployment through automation of white-collor jobs, as predicted by Frey and Osborn in their widely noticed work on \textit{The future of employment} \cite{frey2017future} is connected to societal ethics and not directly to machine ethics and will thus not be the focus of this paper. In contrast, the topic of AI safety, popularized by Nick Bostrom's work on \textit{Superintelligence: Paths, dangers, strategies} \cite{bostrom2014superintelligence} is very relevant to our inquiry. Moral machines should be constructed such that they neither pose a threat to individual humans, nor to humanity as a whole. That this is no easy challenge is illustrated by science fiction stories like the robot tales of Isaac Asimov and especially the issues discussed through his \textit{Three Laws of Robotics}. Our response to the AI safety problem is again to take virtue ethics as the guiding moral theory for building moral machines, as we try to show in this paper. 

Just like in human ethics, one can argue over the correct moral standards for machines, but it should be agreeable that these must ultimately be defined by humans \footnote{Actually, this thesis is not universally accepted. There are also proponents for machines computationally constructing a perfect morality system.}. While there exists no theoretical reason against the possibility of emergent moral-like behavior patterns in a group of autonomously interacting machines, similar to how moral naturalists assert it for human morality, we also exclude the debate of machine-machine morality from this paper and focus on morality in the interaction between humans and robotic systems like the above-mentioned examples. We will not discuss the exact contents of an appropriate machine morality, meaning which specific actions the machine should perform in certain situations, but on a higher abstraction level about the structure of the morality system that needs to be built into the machines. Only an answer to this structural question allows the goal-directed handling of the specific ethical values and norms as their formulation depends on the chosen moral theory. 
While deontology focuses on the formulation of duties and maxims, the objective of a consequentialist ethics is the  identification of desirable and maximizable consequences of actions, e.g. happiness. In virtue ethics, the moral theory we argue for, the moral actor with her character dispositions is put into the center, instead of her moral judgments and actions. Utilizing a seemingly circular phrasing, one can say that in virtue ethics moral actions are conducted by virtuous moral agents. With respect to machines this changes the primary question from "With which algorithm can machines choose the right action?" to \textit{"How can we build a machine that, owing to its constitution, acts appropriately in arbitrary situations?}. Our main thesis answers this guiding question by proposing to take inspiration from virtue ethics when attempting to build moral machines. 


\paragraph{AMAs - Autonomous moral agents} In this paper we consider machines whose autonomy through artificial cognitive capabilities is large enough to allow moral consideration of possible actions. We abstract from the concrete implementation of these machines by using the term \textit{"autonomous moral agent"} as coined by Wendell Wallach \cite{wallach2008moral} as a portmanteau word for robots, AIs and cognitive systems in general.

\section{Virtue Ethics and AI}

In contrast to normative moral theories which focus on the ethical content of actions, virtue ethics deals with the moral agent and her virtues. \\
Virtues are character dispositions which can be obtained through habituation and are thus not limited to moral dimensions. This extension is especially visible in Aristotle's Nicomachean Ethics, on which we will focus in the following. In his theory of the soul, Aristotle introduces dianoetic virtues of the rational mind alongside the ethical virtues and describes how they relate to each other through practical wisdom. 
Aristotle begins his Nicomachean Ethics with the teleological thesis that everything pursues a goal (\textit{telos}) and that every species has a specific function or purpose (\textit{ergon}). This function and thus the goal of all pursuit is not provided externally, but lies within the nature of each living being. A good life means for Aristotle to fulfill one's \textit{ergon} through the species-specific way of life and to thus exhibit virtue (\textit{aret\=e}). This is his answer to the central philosophical question of antiquity about the good life: it is to be found in happiness, \textit{eudaimonia}, and he proclaims that it can be achieved through virtuous behavior, according to one's \textit{ergon}. In the first book of the Nicomachean Ethics Aristotle then undertakes a systematic search for the \textit{ergon} of man. He eventually comes to the conclusion that humans distinguish themselves above all else through their capability for reason, and thus proposes that our \textit{aret\=e} lies in excellent use of the rational part of our soul through the right actions.

\subsection{Teleological model of pursuit - Goal-oriented behavior in cybernetics and AI}
The teleological form of Aristotelian virtue ethics offers an excellent link to modern AI research. While the concepts of teleology and final causes take a fixed, if not central, place in Aristotle's natural philosophy, they have been exiled from scientific vocabulary by Francis Bacon, the founding father of modern science. Not until the middle of the 20th century has teleology been returned from exile by the cyberneticists and been recognized as a principle common in living beings and machines. The title of the founding paper of cybernetics \textit{"Behavior, purpose and teleology"} \cite{rosenblueth1943behavior} by Rosenblueth, Wiener and Bigelow illustrates this reversed paradigm change nicely. If teleology is a common principle in biology and technology, as the cyberneticians proclaim, then it might also be used as an integrator for the morality of both spheres. 

It does not surprise that Norbert Wiener's own ethical position, which he presented in his book \textit{The Human Use of Human Beings} in 1950 \cite{norbert1954human} and in later writings, was strongly Aristotelian in nature, as Terrell W. Bynum recognized \cite{bynum2005norbert}. Bynum argued that this connection was based on their similar natural philosophies (e.g. understanding animals and humans as information-processing beings and deriving their purpose from it) and not only on the general goal-orientation of moral actions like in other teleological moral theories such as consequentialism. 
In a paper from 1960 Wiener already described what contemporary AI ethicists call the \textit{value alignment problem}, according to which a central challenge for building moral machines is to align the values after which an autonomous machines chooses its goals with the complex values of humans, which can hardly be formulated explicitly. Failed attempts to do so can be found in many literary works, from the Greek myth of King Midas to Goethe's Sorcerer's Apprentice. 

\begin{quote}
\textit{"Disastrous results are to be expected not merely in the world of fairy tales but in the real world wherever two agencies essentially foreign to each other are coupled in the attempt to achieve a common purpose. If the communication between these two agencies as to the nature of this purpose is incomplete, it must only be expected that the results of this cooperation will be unsatisfactory. If we use, to achieve our purposes, a mechanical agency with whose operation we cannot efficiently interfere once we have started it, because the action is so fast and irrevocable that we have not the data to intervene before the action is complete, then we had better be quite sure that the purpose put into the machine is the purpose which we really desire and not merely a colorful imitation of it."} \cite{wiener1960some}
\end{quote}

Cybernetics can be seen as a historical and intellectual precursor of artificial intelligence research. While it had strong differences with the cognitivistic GOFAI (good old-fashioned AI), cybernetic ideas are highly influential in modern AI. The currently successful field of artificial neural networks (synonymous terms are connectionism and deep learning) originated from the research of the cyberneticians McCulloch, Pitts and Rosenblatt. 
Goal-directed planning is a central part of modern AI and especially of advanced robotics. In contrast to other forms of machine learning like supervised or unsupervised learning, \textit{reinforcement learning} is concerned with the goal-driven (and therefore teleological) behavior of agents. 

This conceptual parallelism between goal-directed planning and learning and Aristotle's motivation for virtue ethics is a first indication that virtue ethics is an excellent fit for building moral machines. 


\subsection{Experience related habituation - Machine learning}
As mentioned in the last section, reinforcement learning is a subcategory of machine learning, which itself is responsible for much of the recent progress in artificial intelligence. This progress has been mainly fueled by increased computing power and sufficiently big data sets. As machine learning has been successful in several domains of AI such as object recognition and natural language processing, it appears reasonable that it might also be successfully utilized for building moral machines. 
The bottom-up approach of learning moral actions is rather different from the top-down approaches of building a moral expert system by feeding it moral facts and designing it to do logical inference on them (similar to the deontological position) or defining an explicit value function which is maximized by the consequences of the correct action (similar to consequentialism). This new bottom-up approach based on machine learning favors or at least is compatible with several meta-ethical positions, e.g. it allows non-cognitivistic and anti-universalistic positions, and thus needs philosophical justification in the form of an underlying moral theory. In virtue ethics the notions of learning and habituation are central, which is not the case for deontology nor for consequentialism. Therefore, we suggest that an integration of machine learning with virtue ethics will be more seamless and natural than with other moral theories. 
Aristotle writes: 
\begin{quote}
\textit{"[...] a young man of practical wisdom cannot be found. The cause is that such wisdom is concerned not only with universals but with particulars, which become familiar from experience"} (NE 1141b 10)
\end{quote}
Applied to AI ethics this means that a machine cannot have practical wisdom (and thus can't act morally) before it has learned from realistic data. Machine learning is the improvement of a machine's performance of a task through experience and Aristotle's virtue ethics is the improvement of one's virtues through experience. Therefore, if one equates the task performance with virtuous actions, developing a virtue ethics-based machine appears possible.

If one includes the consideration of where the reward function of reinforcement learning comes from into the consideration, one arrives at a hybrid system, with top-down and bottom-up processes similar to how Wendell Wallach sees virtue ethics \cite{wallach2008moral}. In our view, there are three mechanisms that contribute to the shaping of a moral reward function. Firstly, there is the external feedback from the environment. In many situations where we act immoral, other people react negatively towards us, which forms a negative reward. Secondly, there is internal feedback through self-reflection. A good way to improve one's morality is to reflect upon past actions and reason about if the chosen action was the right one or if one could have acted better. Self-scolding resembles negative reward and self-praise resembles positive reward. A third way to shape the moral reward function is through observation of moral exemplars.

\subsection{Soul theory - Cognitive Systems}
A side product of Aristotle's search for the human \textit{ergon} is a detailed insight into his theory of the soul (or, based on the Greek word for soul (\textit{psych\=e}): psychology). He divides the human soul into a vegetative part (\textit{phytik\=on}), a perceptive part (\textit{aisth\=esis}), the desires (\textit{\=orexis}) and reason (\textit{logos}). These parts of the soul interact with each other (e.g. the dianoetic virtue of practical wisdom (\textit{phron\=esis}) controls the desires) and are thus integrated in a cognitive system. The similarities to cognitive systems research become even stronger in the second book of the Nicomachean Ethics, in which Aristotle describes how virtues are acquired.

\begin{quote}
\textit{"Neither by nature, then, nor contrary to nature do the virtues arise
in us; rather we are adapted by nature to receive them, and are made perfect by habit.} (NE 1103a 25)
\end{quote}

This reminds strongly of the distinction of ontogenetic and phylogenetic development of cognitive systems \cite{vernon2014artificial}. According to this position, cognitive systems acquire some of their capabilities and characteristics already during their development (phylogenetic development), especially a capacity of learning which bootstraps the acquisition of many other skills and functions during lifetime, adapted to the specific needs (ontogenetic). 
The popular hypothesis that one could develop an ethics module which can be additively appended to a cognitive system seems improbable within the virtue ethics approach because the moral behavior is a property of the overall system. According to Aristotle, virtue is to be found in the excellent use of our cognitive apparatus and therefore, it does not make sense to view it distinct from other cognitive functions. 
Our position is thus compatible with Joshua Greene's view on moral cognition after which there is no dedicated neural circuit for morality, but it is instead unified at the functional level \cite{greene2015rise}. 

Instead of the ethics module approach, we propose that the development of a moral machine must be conducted in two successive steps. In the first step, one constructs the cognitive architecture of the system, endowed with the innate capability to form character dispositions through learning and habituation. In the second step the system is put into an environment in which it can learn the desired character dispositions through suitable interaction.  

A closer look at the structure of Aristotle's \textit{ergon}-argument allows to break with two common misconceptions which seem to render a virtue ethical approach in machine ethics impossible. The first misconception is ethical anthropocentrism, after which only humans can act morally. This might have been correct in the past, but only because humans have been the only species capable of higher-level cognition, which, according to Aristotle, is a requirement for ethical virtues and thus moral action. If there was another species, for example a machine, with the same capacity for reason and dispositions of character, then it appears probable that its \textit{aret\=e} would also lie in excellent use and improvement of those. The second misconception of Aristotle's virtue ethics is that it takes happiness to be the goal and measure of all actions. Since machines are not capable of genuine feelings of happiness, it is argued, that virtue ethics can't be applied to them. This argument is based on an erroneous understanding of \textit{eudaimonia}. Aristotle does not mean psychological states of happiness nor maximized pleasure, as John Locke defines 'happiness'. The Greek term \textit{eudaimonia} has a much broader meaning and refers mainly to a successful conduct of life (according to one's \textit{ergon}). 
A virtuous machine programed to pursue \textit{eudaimonia} would therefore not be prone to \textit{wireheading}, which is the artificial stimulation of the brain's reward center to experience pleasure.

\textit{Eudaimonia} is thus more than mere subjective perception, it originates from a moral realism. 

\subsection{Lifeworld - Dynamical Interaction with the Environment}

Aristotle is concerned with the lifeworld, the world in which  situated agents live and perceive, which builds a strong contrast to Plato's idealist focus on the beyond. 
This opposition is alluded to in Raphael's \textit{The School of Athens}, where the Renaissance artist painted both philosophers in the center, Plato pointing towards the sky (the beyond) while Aristotle, holding his \textit{Nicomachean Ethics} in hand, points downwards at the earthly realm. 
In ethics this disparity is visible in Aristotle's emphasis of \textit{phron\=esis} over Plato's \textit{sophia}. Aristotle lays special focus on the unique characteristics of every situation that needs moral action instead of calling for universal rules and duties. 

\begin{quote}
\textit{"Nor is practical wisdom concerned with universals only - it must also recognize the particulars; for it is practical, and practice is concerned with particulars."} (NE 1141b 15)
\end{quote}

The field of artificial intelligence is faced with a similar clash. It's dispute is based on the question whether intelligence requires a body. While the field of embodied cognition has been dominant in the 1990s and early 2000s, disembodied approaches to AI have recently had major successes, especially in the AI-subfield of machine learning. 

Out of the three subcategories of machine learning, supervised learning, unsupervised learning and reinforcement learning (RL), the latter is the lifeworldly approach. In contrast to the other two, RL is based on dynamic interaction with the environment, of which the agent typically has only imperfect knowledge. Reinforcement learning divides the world into two parts, the agent and the environment. The agent takes actions based on its knowledge of the environment's current state. This action changes (updates) the environment's state and elicits a reward feedback which the agent receives to improve the policy with which it chooses actions based on environment states. This improvement is called learning. 

Reinforcement learning could also be connected with the moral theory of utilitarianism, a variety of consequentialism, since both have the aim of maximizing estimated utility. 
However this combination is not intuitive since utilitarianism does not provide for a learning process, which is central to reinforcement learning.  

Virtue ethics incorporates both, the value-maximizing policy and the focus on learning. 

\begin{quote}
\textit{"The man who is without qualification good at deliberating is the man who is capable of aiming in accordance with calculation at the best for man of things attainable by action."} (NE 1141b 10)
\end{quote}
This quote from Aristotle shows that a virtuous person is capable of deliberately choosing the action which promises the best outcome for man. Besides taking a position opposite of ethical egoism by paying attention to 'man' instead of exclusively oneself (similar to utilitarianisms slogan 'the greatest happiness for the greatest number'), the quote indicates that this deliberate process can be calculated (and could thus be computed by a machine). In contrast to utilitarianism, the 'best', which is aimed for, is not utility, but what is best for the good life, \textit{eudaimonia}. This difference is important as it determines the definition of the \textit{reward signal} of the virtue ethics based reinforcement learning system. For biological systems this reward is often specified as either pleasure or pain, which fits well with hedonistic utilitarianism. In virtue ethics the  reward system is much more complex. Pleasure and pain also exist in animals and can therefore not be a major part of man's \textit{ergon}. But it is this \textit{ergon} which, according to Aristotle, needs to be optimized through improving it until excellence (\textit{aret\=e}). Only this process can lead to genuine happiness (\textit{eudaimonia}) and thus a good life.  

The major challenge for the combination of virtue ethics and reinforcement learning lies hence in the exact specification of an \textit{eudaimonic} reward signal. 
In the next chapter we will take a closer look on several virtues from which we will derive an approach for solving this  problem.



\section{Virtues of a Cognitive Machine}

In his \textit{Nicomachean Ethics} Aristotle takes up Plato's four main virtues, which his teacher presented in his dialogues \textit{Politeia} and \textit{Nomoi} and which were named cardinal virtues by Thomas Aquinas in the 13th century. 
According to the medieval church father, all other virtues are hanging on these cardinal virtues of \textit{prudence, courage, temperance} and \textit{justice}, similar to how a door hangs on its hinge. The reception and adaptation of Aristotle's ethics and his other writings through Thomas Aquinas shaped the Christian doctrine and life through the Middle Ages and can still be found in the catechism of the Catholic church. 
Starting with prudence, we will present the four cardinal virtues according to their description in the \textit{Nicomachean Ethics} and will discuss their application on autonomous moral agents. 
Aristotle went further than Plato and discussed several virtues of character beyond the cardinal virtues, coming from all areas of life, from financial generosity through truthfulness to friendship. This life-proximity is characteristic of Aristotle's ethics and renders it a truly practical philosophy. 
While other moral theories argue in idealistic scenarios and thought experiments (and are partly only valid in those), Aristotle's virtue ethics does not shy away from the messy, uncertain and diversity of the real world. Instead of raising universal rules, after which everyone ought to act, there are only virtues of character which describe a general striving through desires. The appropriate action in each particular case is produced only in combination with prudence, which Aristotle calls \textit{phron\=esis}.

\subsection{Prudence/ Practical Wisdom (\textit{phron\=esis})}

The first of the four main virtues which Aristotle takes up from Plato is prudence, often also called by its more descriptive name practical wisdom. In contrast to the other virtues, it isn't an ethical but a dianoetic virtue as it relates to reason instead of character. This partition originated in Aristotle's soul theory in which he lists virtues of reason (dianoetic virtues) next to virtues of character (ethical virtues) as properties of the intelligent part of the soul. The virtues of reason comprise the virtues of pure reason and the virtues of practical reason. Pure reason includes science (\textit{epist\=em\=e}), wisdom (\textit{sophia}) and intuitive thought (\textit{no\=us}). Practical reason on the other hand refers to the virtues of craftsmanship (\textit{techn\=e}), of making (\textit{poi\=esis}) and practical wisdom (\textit{phron\=esis}). According to this subdivision in pure and practical reason, there exist two ways to lead a good life in the eudaimonic sense: the theoretical life and the practical life. AI systems can lead a theoretical life of contemplation, e.g. when they are applied to scientific data analysis, but to lead a practical life they need the capacity for practical wisdom and morality. This distinction in theoretical and practical life of an AI somewhat resembles the distinction into narrow and general AIs, where narrow AI describes artificial intelligence systems that are focused on performing one specific task (e.g. image classification) while general AI can operate in more general and realistic situations. A comparison to weak and strong AI is less fitting as this distinction focuses on consciousness, which Aristotle's division in theoretical and practical life does not. In the following we will focus on machines that take part in the practical life.
According to Aristotle, practical wisdom can not be taught but has to be learned through realistic experience. For machines this means that practical wisdom can't be preprogrammed into them but has to come as a result of machine learning. As shown in the last chapter this is because practical wisdom is concerned with particulars, meaning with specific real-world situations instead of abstracted general scenarios. In Aristotle's opinion it is therefore possible that an experienced person acts better than a knowing person. 

\begin{quote}
\textit{"Now practical wisdom is concerned with action; therefore one should have both forms of it [knowledge about universals and particulars], or the latter in preference to the former. But here, too, there must be a controlling kind [of practical wisdom]."} (NE 1141b 20)
\end{quote}

This is where Aristotle's virtue ethics strongly diverges from  other moral theories. Kant's deontological duty ethics concerns itself explicitly with universals as illustrated in his categorical imperative: \textit{"Act only according to that maxim whereby you can at the same time will that it should become a universal law"} (Immanuel Kant, Grounding for the Metaphysics of Morals, AA IV, 421). For the development of moral machines Aristotle's position entails that morality can't be programmed in as declarative general laws and rules, but must be learned through the experience of interaction with one's environment. This experience must come from reality itself, otherwise it is useless. Abstract, unrealistic and bookish thought experiments like the \textit{trolley problem} are irrelevant for Aristotle. First of all it is a highly improbable scenario. It seems unlikely that there has ever been a situation where there were five people tied up on one railway track and one person on a second railway track with another person standing next to a lever for switching an approaching trolley's direction. 
One might argue that this scenario is only an abstraction which can be shaped into different more specific scenarios, e.g. of an autonomous car having to decide whom to kill in case of an inevitable crash. A virtue ethicist would respond to this that it is then a different situation and thus in need of different moral consideration. For them it is exactly the specifics of each concrete situation that constitute the moral consideration. 
Another problem with the trolley problem is its static nature. There is no dynamic interaction with the environment and no feedback to one's actions. 
A third issue that a virtue ethicist takes with the trolley problem is that its moral situation is pre-given. As we will explain in more detail when we turn towards \textit{moral attention} in the next section, one major task of moral action is to detect the moral situation in the first place.

The philosopher of technology Shannon Vallor describes the lifeworldly nature of practical wisdom in her work on \textit{Technology and the Virtues} very fittingly: 

\begin{quote}
\textit{"Practical wisdom is often classified as an intellectual virtue because it involves cognition and judgment; yet it operates within the moral realm, uniting cognitive, perceptual, affective, and motor capacities in refined and fluid expressions of moral excellence that respond appropriately and intelligently to the ethical calls of particular situations."} \cite[p. 99]{vallor2016technology}
\end{quote}

Virtuous practical wisdom is an interactive and lifeworldly disposition and therefore involves the whole situated agent. The process of developing a moral machine can not be reduced to programming a rational decision-system, as it won't have the ability to perceive its environment nor to act on its own. Instead, we must lay our focus on the whole cognitive system, including the perceiving and acting parts. In other words: following a virtue ethical position, morality needs a body. 
Thus Aristotle's ethics and psychological views at large have strong commonalities with the modern paradigm of \textit{embodied cognition} (see \cite{pfeifer2006body}). 
Manners and ways of behavior learned in one culture can not always be transfered to a different culture. The same applies for moral behavior. There are actions perceived as morally correct in some cultures that are unacceptable in others. In AI research this discrepancy between the environment in which the learning process took place and the environment in which the learned models are used is called \textit{distributional shift} between \textit{training data} and \textit{test data}. 

To minimize the \textit{distributional shift}, an AMA would have to learn in the real world. This however poses a problem, because it is important to experience as many different situations as possible during the training phase, in other words to get a good sampling of the data. Otherwise, the learned model will not generalize well and only work in those specific kinds of situations. Especially the training of appropriate actions in morally extreme situations is of high importance for the security of these systems. As the name indicates, morally extreme situations occur extremely scarcely, which means that an AMA learning in the real world would have to train for a very long time to converge to a stable policy for those rare scenarios. One could, of course, produce these situations deliberately, but this would be morally questionable as it is in the nature of morally extreme situations that they involve dilemmata in which there cannot only be winners and loosing is harmful. Re-enacting these situations would also be an imperfect solution as subtracting the possibility of danger would decrease the realism of the situation and thus produce a distributional shift. Realistic simulations of morally relevant situations might still be the best option of training AMAs without getting people hurt. 
Reinforcement learning involves both \textit{exploration} and \textit{exploitation}, meaning  a trade-off between actions with the goal of discovering new states and testing alternative actions and performing actions that according to past experience promise the highest reward. An AMA that purposely explores radically new actions has the potential to bring about highly undesirable consequences. If they are to be trained in the real world (or physical re-enactments thereof) it is important to be able to impose safety constraints upon the field of possible actions. 

In her work on the relationship between technology and the virtues, Shannon Vallor introduced several requirements for practical wisdom \cite[p. 99ff]{vallor2016technology}. According to her practical wisdom demands three abilities: moral attention, an extension of moral concern and prudential judgment. 

\paragraph{Moral Attention}
In the fields of cognitive science and artificial cognitive systems the topic of visual attention is highly researched. It is necessary for physical agents to focus their perception as otherwise the inflow of information would be too large for sufficiently fast reactions. In real situations making decisions and acting according to them must happen in real-time. This applies not only for biological agents, but also for AMAs. When an autonomous vehicle detects a ball rolling onto the street, it is morally insufficient if the AMA takes an hour for the ethical consideration by taking all possible consequences of each possible action into account in a consequentialist fashion. Lifeworldly morality therefore not only needs situation-appropriate compromises between reaction time and time for moral consideration, but also has to be systemically designed such that it allows for fast reaction times. For this reason it needs mechanism for information reduction and selection, e.g. moral attention. 
In the lifeworld the moral situations have to be found in the first place. In contrast to thought experiments AMAs in the real world are not told that a moral significant situation is coming up, nor is it told which information will be relevant and which possible actions it has available. An AMA has to detect these situation itself, e.g. through \textit{unsupervised learning} by clustering situation into those who require moral action and into those that don't. An example is taking the bus which usually is not a morally relevant action. However, once an elderly lady steps in and all seats are already taken, morality requires to offer one's seat. This is of course a morally relevant situation, but the ethical decision process, the moral judgment, is not the main challenge. Instead, the main challenge is being attentive to one's environment so that one can perceive and recognize such situations. This attentiveness is a trait of character, one that can be learned through habituation, a virtue if you will. I argue that many, if not most of our real-world moral situations are actually non-obvious but have to be detected. Machine morality thus has to be more than a decision system, but has to incorporate other cognitive functions as well. 

And, as a side note, this is where our relation with technology comes into play as it shapes our own attentiveness. Many people nowadays have the habit of pulling out their smartphones as soon as they embark public transportation. This, of course, is not morally wrong, but we have to be aware that it changes our attentiveness and thus our ability to recognize morally significant situations. But we shouldn't be technophobic. If technology can shape our moral attention for the worse, it certainly has the capability to shape it for the better as well. 

In the language of visual attention we could say that the embarking of the elderly lady produces a moral saliency that adds up with other salience features to produce a moral saliency map which can guide one's moral attention. However, these saliences are not all absolute, but some are relative, e.g. the moral saliency of the elderly lady increases if there are no free seats. An AMA would have to know these relations and be programmed so that it checks them. 
In robotics there are two main approaches of tackling the challenge of visual attention in machines: bottom-up and top-down. The bottom-up approach is based on visual salient features like color (red is very salient) and orientation (an object oriented differently than all objects around it is very salient). These features are weighted and summed up to produce the saliency map. 
In contrast, the top-down approach is based on huge datasets of images and respective points of attention where human subjects were focusing on, acquired through gaze detection systems. Using \textit{supervised learning} algorithms, e.g. deep neural networks, one can then learn a mapping from the images to the points of attention, which are then taken as points of highest saliency. 

We propose that similar approaches can be taken in moral attention systems as well. We believe that emotions play a major role for the bottum-up approach. Partly, this has to do with faster computation (remember that moral action in the real world needs to be real-time). Instead of having to think about why the elderly lady is more needy than other passengers, an emotion of care helps one come to the same conclusion, but faster and with less energy spent. These emotions as well as other features of moral saliency can be either innate or learned. For Aristotle the emotions and affects, especially empathy, play a major role in perceiving moral situations.
Emotions are much more than moral saliences, they additionally build up the urge for action (in the sense of an action potential) and thus bridge the gap between perception and action. When we see a needy person or a child looking for help we don't just objectively perceive the situation but are pushed towards helping. 

All situations possess a certain moral relevance, but the saliency is high enough in only in a few of them to require moral action. 

In a top-down approach, recognition of morally relevant situations can also be learned through experience, e.g. through rewards (if you stand up once the lady steps into the bus and you get praise or a thank you) or through observing moral exemplars. Aristotle writes about this: 
\begin{quote}
\textit{"Therefore [age brings intuitive reason and judgment] we ought to attend to the undemonstrated sayings and opinions of experienced and older people or of people of practical wisdom not less than to demonstrations; for because experience has given them an eye they see aright.} (NE 1143b 10)
\end{quote}

The main disadvantage of the top-down approach to moral attention is that it is hardly possible to give reasons for the result, just like in top-down visual attention systems. 

\paragraph{Extension of moral concern}
Vallor writes that another challenge of moral self-cultivation lies in an appropriate extension of moral concern. This extensions should be \textit{"to the right beings, at the right time, to the right degree, and in the right manner"} \cite[p. 110]{vallor2016technology}. Modern forms of utilitarianism are often based on moral universality and indifference. According to these principles, needs of people in foreign countries are to be weighted equally to the needs of people in the moral vicinity when calculating the overall utility. This universalism does not exist in virtue ethics, but instead it calls for an extension of moral concern that is appropriate for the specific situation. Here is where the lifeworldlyness of virtue ethics comes into play again, as it is simply impossible to take into account the needs of every human (and animal) for every action. For certain actions the moral relevant area has to be extended to all of humanity, offering a connection to Hans Jonas' ethics of responsibility. This connection serves as a reminder that the extension of moral concern has to be not only in space but also in time due to the power of our technologies. 
For an AMA the capability for dynamic and appropriate extension of moral concern is of high importance. A utilitarian service robot which has been programmed with ethical equality of all humans would probably rather quickly move to an area of crisis where its help would lead to much higher utility than in the luxury apartment of the people that are most likely to be its first buyers. 
While Aristotle does not exclude the value of considering the moral far field, he poses the focus of his virtue ethics on the moral near field, which becomes especially evident in his inquiry about friendship which we will discuss later.

\paragraph{Prudential judgment}
According to Vallor, the third dimension of practical wisdom is prudential judgment which chooses the appropriate action for the given situation. Prudential judgment puts each virtue of character into its mean (\textit{mesot\=es}) and thus away from the extremes. The virtue of courage for example lies in the mean between foolhardiness and cowardice.  

As shown above, prudential judgment can't be taught, but has to be learned from experience. This is possible in two ways. Firstly, through direct interaction with the environment and subsequent reflection and secondly through observing the behavior of morally excellent people which are called \textit{moral exemplars} or \textit{moral saints}. 
In order to build a prudent AMA one can follow these two approaches. The first approach of interactive learning through experience can be implemented by \textit{reinforcement learning}. Dependent on the situation (which is perceived through moral attention) the agent interacts with its environment and tests different possible actions. Through the subsequent reflection about whether the prior goal was meet, a reward signal is generated, allowing the AMA to learn a policy that maximizes the expected return (sum of all future discounted reward). 
The second approach is based on imitation learning from exemplars which is an active research topic in robotics, AI and neuroscience. In artificial intelligence research one method of imitation learning is \textit{inverse reinforcement learning} (IRL) \cite{ng2000algorithms}. While forward RL generates an optimal policy given a reward function, inverse RL turns the learning process around and learns an exemplar's most probable reward function based on observation of its behavior, which is a proxy of its policy. The estimated reward function can then be used in forward reinforcement learning through which the agent learns a policy of its own which resembles the original policy of the exemplar. This connection of IRL and RL is called \textit{apprenticeship learning} \cite{abbeel2004apprenticeship} and plays an important role in contemporary AI safety research. 
For the realization of \textit{apprenticeship learning} for AMAs it is necessary to produce usable data sets of behavioral execution traces from moral exemplars in many different realistic situations. One possible starting point could be depictions of actions from historic exemplars, like in \textit{Saints and Virtues} \cite{hawley1987saints} or even more classical in antique writings like those of Plutarch or religious ones like the bible and reports about Muhammad. Helping people learn to become more virtuous has been and still is a major purpose of these texts. Thus it doesn't seem too far-fetched that they might also be used for the moral education of machines in the future. Every cultural zone could then select (democratically) the exemplars after which its cognitive machines should be built. 
However, the available datasets of these historic moral exemplars are most probably too small to allow for robust behavioral imitation learning and furthermore not suitable as the described situations are in many ways different to our situations today. Instead, one would have to build a modern 'Plutarch', for example through crowd-sourcing of behavioral trajectories on different benchmark situations. These individual execution traces would be as good as individual people are, and thus not consistently morally optimal. There is good reason, however, to expect that the moral behavior of the subjects would be better than how they act in real life due to them knowing that their behavior is being recorded. 
The connection of IRL to AI safety research is illustrated in the paper \textit{Concrete Problems in AI Safety} \cite{amodei2016concrete} in which \textit{reward hacking} is discussed. Reward hacking happens when an RL-based AI system finds ways to maximize its reward in ways that conflict with the original intentions of the developers. The reason for this is that the (moral) intentions of the developers exist only informally (and possible not even inherently consistent). To be used as a reward function for building a moral machine using RL it needs to be formulated explicitly. Usually, one uses substitutive features which approximately correlate with the informal goals. The IRL method allows to circumvent the problem of explicitly defining the developers' moral intentions and thus reduces the dangerous discrepancy that can lead to reward hacking. 

\subsection{Justice (\textit{dikaiosyn\=e})}
Aristotle puts special emphasis on the virtue of justice (\textit{dikaiosyn\=e}) by writing that \textit{"[...] justice is often thought to be the greatest of virtues, and 'neither evening nor morning star' is so wonderful; and proverbially 'in justice is every virtue comprehended'"} (NE 1129b 25). Especially important is that justice is about good actions for others: \begin{quote}
\textit{"For this same reason justice, alone of the virtues, is thought to be 'another's good', because it is related to another; for it does what is advantageous to another, either a ruler or a co-partner."} (NE 1130a).  
\end{quote}
This view is important as it counters the common notion of virtue ethics being egocentric. Machines should be useful to us and therefore it should act towards our good. Thus the virtue of justice takes up a central place in virtue ethics for AMAs as well. 
Aristotle goes further in saying that 
\begin{quote}
\textit{[T]he law bids us practice every virtue and forbids us to practice any vice. An the things that tend to produce virtue taken as a whole are those of the acts prescribed by the law which have been prescribed with a view to eduction for a common good."} (NE 1130b 20)
\end{quote}
In a just life it is included not to commit adultery nor acts of violence, which both require the virtue of temperance. One also has to be  gentle in order to life according to the law so that one does not strike back after every provocation. According to Aristotle, the law prescribes acting courageously and not to flee cowardly when the community is in danger. These examples show how other virtues of character emerge out of justice and form as a whole the basis for our laws. 
Following the laws is only one half of the virtue of justice, the other being fairness.

\paragraph{Acting according to the law}
Any autonomous moral machine should act according to the normative law. In virtue ethics these actions happen with respect to the particular situation and by applying practical wisdom. There are, however, situations in which laws are contradictory and lead to paralysis of the AMA or in which they disagree with common sense. If an autonomous car is still standing behind a red traffic light after three hours, then there has not been virtuous action nor judgment according to the reasonable part of the soul. This example illustrates that actions according to the virtue of justice are not binary, right or wrong, but rather they are also, like the other virtues, on a spectrum between extremes. In justice these extremes are unlawfulness and blind law-abiding. Thus, justice requires practical wisdom (\textit{phron\=esis}) to find the right action according to the particular situation as the mean (\textit{mesot\=es})  between these extremes. 
If an AMA breaks a law (e.g. if it chooses to cross the red traffic light after waiting very long) the question of responsibility poses itself. This is an important question that needs to be addressed, but we will not focus on it in this paper.

\paragraph{Distributive justice and fairness}
For Aristotle a person who wants more than appropriate is an unjust person. The dispositions of wanting too-much or too-little are also poles between which the mean (\textit{mesot\=es}) of justice lies (see NE 1131a). Thus one can describe the virtue of justice as lying in the center of a multidimensional space of dispositions with lawfulness and fairness being its primary axes. 
Aristotle presents two kinds of fair equality. There is equality in the \textit{"distributions of honour or money or the other things that fall to be divided among those who have a share in the constitution"} (NE 1130b 30) and there is equality of transactions between persons. 
Aristotle does not envision an equal distribution of goods in the communist sense. First of all, only public goods are to be divided equally (meaning that there are also private goods to which equality does not apply) and secondly he points out that the distribution should be according to merit, which is defined differently by different people. For democrats the merit is the status of freeman, for others its wealth or noble birth. For an AMA one can implement the merit as a weighting of features which has been either predefined during development or is being learned during operation, e.g. according to cultural preferences. Since an AMA does not need personal belongings it appears reasonable to set its own merit rather low and the one of its owners rather high, though there are also proposals that wealth produced by autonomous machines should be given to the community. A just and fair AMA with an inclination to set its own merit relatively low will not pursue money or power and since virtue will be part of its nature, this disposition will be natural to it with no desire to be changed. Thus, a virtuous AI would not be prone to dystopias in which an artificial intelligence takes over the world. 
As noted before, the second kind of equality concerns equal transactions between persons. In the case of autonomous machines this definition would have to be change to transactions between humans and AMAs as well as between AMAs. An AMA shouldn't rip people off, independent of the prohibition by law. This would be especially important for AMAs utilized for fast-speed trading in financial markets. 
Additional to these volitional transactions in which both parts are aware of the trade and agree upon it, Aristotle also discusses undesired transactions such as theft, adultery, poisoning (all done in secret) and homicide, robbery and rape (all including violence). Of course no AMA should ever commit these crimes. Not only because they are prohibited by law but also because they are unfair transactions. It is important to Aristotle to point out that these actions would not become fair by mere reciprocity in the sense of \textit{"eye for an eye"} as the Pythagoreans thought. A mistreatment is not made equal and thus just by letting the victim mistreat the offender. This is because the circumstances and thus the particulars of the two acts can never be the same, e.g.  \textit{"there is a great difference between a voluntary and an involuntary act"} (NE 1132b 20). 

\subsection{Temperance (\textit{s\=ophrosyn\=e})}
For Aristotle the virtue of temperance (\textit{s\=ophrosyn\=e} in Greek; \textit{temperantia} in Latin) is about bodily pleasures (see NE 1118a). For AMAs this is similar to getting rewards, though the analogy does not hold perfectly because we have defined the rewards of an AMA to be \textit{eudaimonia}, which is more than bodily pleasure. Still, we believe that the analogy can be useful for the design of moral machines and will utilize a broader form of temperance than Aristotle did, beyond bodily pleasures. 
Going with this analogy we have to ask: why would it make sense to act temperate with regards to the accumulation of rewards whose optimization is the primal goal of RL agents? Aristotle sees intemperance as a desire that exceeds the appropriate. To decide what is appropriate and what isn't, one needs practical wisdom and this should also apply to AMAs which try to optimize a reward through reinforcement learning. Reward functions are, especially in the moral domain, extremely complex and cannot be formulated explicitly. Instead they have to be approximated, for example by a quantifiable substitutive feature which correlates with the implicitly known objective quantity. As discussed before, this discrepancy, which is often called \textit{value alignment problem}, can lead to unwanted behavior like reward hacking in which the agent finds ways (hacks) that boundlessly maximize the reward but do not confirm with the developers' implicit expectations. This situation comes about due to an unfortunate combination of never being able to get the correct moral reward function (thus having to work with approximations and replacement measures) and \textit{Goodhart's law}. According to this adage, a measure ceases to be a good measure when it is optimized as a target. One example of this are school grades. They serve as measures for the student's level of education, but if the student exclusively tries to optimize them to the extreme, they loose their value, because education is more than what grades can measure and it is this difference that gets lost by optimization. We propose therefore that AMAs should always be uncertain about their reward function to avoid attempts of gaming it.

Aristotle writes: 
\textit{"[...] in an irrational being the desire for pleasure is insatiable"} (NE 119b 5). This is exactly the fear of a boundless superintelligence with the ability and desire for exponential self-improvement. 

Aristotle's solution to the problem of insatiability is the application of practical wisdom to the virtue of temperance:
\begin{quote}
\textit{"Hence they [the appetites] should be moderate and few, and should in no way oppose the rational principle - and this is what we call an obedient and chastened state - and as the child should live according to the direction of his tutor, so the appetite element should live according to reason."} (NE 1119b 10)
\end{quote}

This is exactly how we want an AMA to behave, like a child that lives according to the directions given by us, its tutors. Since the virtues are intrinsic and inseparable parts of one's character there is additionally no reason for an AMA to change the mechanisms that enforce its obedience. Not even a superintelligence would want to change them as its \textit{eudaimonia} would lie in actions according to its virtues, including temperance. This is important because other approaches to contain a superintelligence lead to the aporia that it could just rewrite its own code and with it all barriers and security confinements that its developers have implemented.


Therefore, we see in the virtue of temperance a prime candidate for solving the \textit{control problem} of a possible future superintelligence, which might in the excess of its insatiability pose an existential threat to humanity. 

Aristotle's final sentence about temperance could thus be extended to: 
\begin{quote}
\textit{"the temperate man [, woman or AMA] craves for the things he ought, as he ought, and when he ought; and this is what reason directs."} (NE 1119b 15)
\end{quote}

This means that a temperate AI would never have the desire for limitless self-improvement as this would constitute an excess. Furthermore, it would not want to change its virtue of temperance as it is part of its character, its very nature. Although it might not seem as such, this is a decisive difference to thought experiments about restricting the AI through additional layers of code, which it might easily rewrite if it is powerful enough and reason dictates that being even more powerful makes achieving its goals or tasks easier. Virtue ethics is not only about reason alone, but about the harmonious interplay of reason and desires: 

\begin{quote}
\textit{"The virtuous agent acts effortlessly, perceives the right reason, has the harmonious right desire, and has an inner state of virtue that flows smoothly into action. The virtuous agent can act as an exemplar of virtue to others."} \cite{iep_virtueEthics}
\end{quote}

\subsection{Courage (\textit{andreia})}
For Aristotle the virtue of courage (\textit{andreia}) lies in the middle (\textit{mesot\=es}) between cowardice, the extreme of fear, and foolhardiness, the extreme of confidence. Courage to him is about the appropriate assessment of risks. Of course, an easily frightened AMA would be rather useless due to it being paralyzed in the face of the many possible negative consequences it would be able to conceive and calculate. It would be like Marvin, the depressed robot from \textit{The Hitchhiker's Guide to the Galaxy}. An AMA that is excessively courageous and thus foolhardy will act wrong in the other extreme, by underestimating risks, not doing anything about them and thus ignoring dangers. 
The answer here too is practical wisdom (\textit{phron\=esis}). It helps to deliberate which fears (risks) are reasonable and can be counteracted, hence allowing courageous actions. Courageous actions is about holding one's ground against what is fearful for the sake of what is valuable: 
\begin{quote}
\textit{"[...] while he [the brave man] will fear even the things that are not beyond human strength, he will face them as he ought and as reason directs, for the sake of the noble; for this is the end of virtue."} (NE 1115b 10)
\end{quote}

Courageous action is not found in the absence of fear, but when one knows about the risks involved and still does what he or she deems best. 
Taking this explanation of courage, we believe that an AMA should be courageous as it should calculate risks and act appropriately upon them. If it observes a human being threatened it should not be afraid of possible harm to itself, but instead of possible harm to the human and therefore intervene. This intervention should be appropriate to the situation, neither too fierce nor to soft. 

Aristotle writes further that \textit{"we show courage in situations where there is the opportunity of showing prowess or where death is noble"} (NE 1115b). Dystopias about a robot uprising often imagine that robots will someday resist against being turned off. If we equate turning off an AMA with death, then this danger does not exist with virtuously courageous AMAs as they would be prepared to face a noble death, i.e. letting themselves be turned off if necessary. It might oppose being turned off for arbitrary reasons, but we don't see any harm (rather some value) in a balanced instinct for self preservation. 
Actually, the virtue of courage conforms with Asimov's third \textit{law of robotics} after which a robot must protect its own existence unless this would conflict with the first or second law (after which he mustn't let humans come to harm and has to obey them). The difference is, that the act of obedience does not have to happen without questioning. After Asimov's laws, a robot would have to terminally turn itself off if a child demanded such on a whim. A courageous AMA wouldn't let that happen as easily. It would demand justification in form of reasons why this action is valuable. For robot systems in certain fields of application where quick turning off might be necessary, e.g. in the military, it would be conceivable to pre-program into the AMA that following orders from specific people is of high value and should be followed immediately. A soldier knows the value of his or her profession and therefore does not have to question every single order. However, history has taught us to be extremely careful with such blind obedience. 

\subsection{Gentleness (\textit{praot\=es})}
An AMA whose function (\textit{ergon}) it is to communicate and interact with humans should have emotions. These play an important role in human interaction and their implications on our pitch and modulation of speaking, facial expressions and chosen vocabulary represents a considerable part of the transmitted information during communication. In order to serve the function of interacting with humans optimally, and therefore achieve excellence in the sense of \textit{ar\=ete}, it is necessary for AMAs to read, interpret and express emotions. The scientific field which concerns itself with these challenges is called \textit{affective computing} and occasionally also \textit{artificial emotional intelligence}. According to the theory of Paul Ekman the expression of emotion in human faces is determined naturally, independent of culture. Therefore, one could classify emotions in the universal categories, e.g. anger, disgust, surprise, happiness. A category which is not explicitly mentioned, but which is of importance to Aristotle, is gentleness (\textit{praot\=es}), which in some translations is also called 'good temper'. One can define it as the mean between irascibility and emotional passivity. An AMA that is used as a care robot in an elderly care home should not be passively going about its duties, but also exhibit gentle and empathic behavior. An AMA deployed as an intelligent tutoring system should not be easy to anger so that it doesn't intimidate its students. But it should also not let every type of student behavior slide without some angry scolding, because otherwise it won't receive any respect and thus couldn't perform its specific function (\textit{ergon}) with excellence to acquire virtue (\textit{ar\=ete}).

Contrary to the frequent opinion that a moral machine should always be happy and friendly, we believe that it should be gentle and of good temper and therefore being able to appropriately act in the face of inappropriate actions towards it. Amongst other reasons, this has to do with acceptance. 

Just like all the other virtues, the virtue of gentleness also requires practical wisdom because it has to define \textit{"how, with whom, at what, and how long one should be angry"} (NE 1126a 30). 

\subsection{Friendship to humans (\textit{philanthr\=opos})}
Aristotle writes in his \textit{Nicomachean Ethics}:
\begin{quote}
\textit{"[P]arent seems by nature to feel it [friendship] for offspring and offspring for parent."} (NE 1155a 15)
\end{quote}

Since artificial agents of any kind are our collective (or at least their developers') offspring, our \textit{mind children} \cite{moravec1988mind}, one could expect that they would be friendly towards humans, without explicitly programming it in. It is, however, questionable whether there can ever be true friendship between man and machine, due to their categorical differences. Aristotle explains that it is controversial whether harmony emerges out of differences, like Heraklit says, or if Empedokles is right in saying that like aims at like (NE 1155b 5). Furthermore, Aristotle writes that love to lifeless objects cannot be regarded as friendship. It is, however, conceivable that his understanding of living things does not exclusively refer to carbon-based biological life as we would define it today, but more generally to active and behaving entities. It is undisputed that true friendship is bidirectional, meaning that it has to originate from both parties. Aristotle believes that in order to receive friendship one has to be likable and therefore pleasant and useful. AMAs are machines and therefore per definition useful (at least for some people, because otherwise they would not have been built or are not working correctly). Machines that work perfectly can also be regarded as pleasant by their users. These two attributes, useful and pleasant, are not enough for perfect friendship. This highest form of friendship can only occur between persons that are good and equal in virtue. \textit{"Now those who wish well to their friends for their sake are most truly friends"} (NE 1156b).
Only a truly virtuous machine could therefore have a chance of building a perfect friendship with a human, and that only if it could be capable of wishing someone well for their sake. 

Friendship to humans is connected to another virtuous disposition, friendliness, which Aristotle discusses separately (see NE 1126b 10). In contrast to friendship, friendliness solely originates from the actor and is equal towards familiar and non-familiar persons. We believe that AMAs need both dispositions. 

\section{Challenges}

Virtue ethics is not without challenges, and these challenges also exist and might even be amplified for virtuous AMAs. 
Here, we want to focus on the most pressing challenge for virtue-based (but also any other) moral machines: \textit{responsibility} and the ability to give reasons for one's actions. 

In contrast to deontology and consequentialism, virtue ethics has a hard time giving reasons for its actions (they certainly exist, but are hard to codify). While deontologists can point towards the principles and duties which have guided their actions, a consequentialist can explain why her actions have lead to the best consequences. An AMA based on virtue ethics on the other hand would have to show how its virtues, which gave rise to its actions, have been formed through experience. This poses an even greater problem if its capability to learn virtues has been implemented as an artificial neural network, due to it being almost impossible to extract intuitively understandable reasons from the many network weights. In this instance, the similarity between virtue ethics and machine learning is disadvantageous. 
Without being able to give reasons to one's actions, one cannot take over responsibility, which is a concept underlying not only our insurance system but also our justice system. If the actions of an AMA produce harm then someone has to take responsibility for it and the victims have a \textit{right to explanation}. The latter has recently (May 2018) been codified by the EU General Data Protection Regulation (GDRP) with regards to all algorithmic decisions. 

\section{Summary}

Condensed to the most important ideas, this work has shown that 
\begin{enumerate}
\item Virtue ethics fits nicely with modern artificial intelligence research and is a promising moral theory as basis for the field of AI ethics.
\item Taking the virtue ethics route to building moral machines allows for a much broader approach than simple decision-theoretic judgment of possible actions. Instead it takes other cognitive functions into account like attention, emotions, learning and actions.
\end{enumerate}

Furthermore, by discussing several virtues in detail, we showed that virtue ethics is a promising moral theory for solving the two major challenges of contemporary AI safety research, the \textit{control problem} and the \textit{value alignment problem}. A machine endowed with the virtue of temperance would not have any desire for excess of any kind, not even for exponential self-improvement, which might lead to a superintelligence posing an existential risk for humanity. Since virtues are an integral part of one's character, the AI would not have the desire of changing its virtue of temperance. 
Learning from virtuous exemplars has been a process of aligning values for centuries (and possibly for all of human history), thus building artificial systems with the same imitation learning capability appears to be a reasonable approach.

\bibliographystyle{agsm}
\bibliography{mybib}{}

\end{document}